\documentclass[11pt]{article}
\usepackage{geometry}                
\usepackage{graphicx}
\usepackage{amssymb}
\usepackage{epstopdf}

\usepackage[utf8]{inputenc} 
\usepackage[T1]{fontenc}    
\usepackage{hyperref}       
\usepackage{url}            
\usepackage{amsfonts}       
\usepackage{amsmath}

\title{Residual Networks as Geodesic Flows of Diffeomorphisms}

\author{
  François~Rousseau\\
  Institut Mines Télécom Atlantique\\
  LaTIM U1101 INSERM, UBL\\
  Brest, France \\
  \texttt{francois.rousseau@imt-atlantique.fr} \\
  \and
  Ronan~Fablet\\
  Institut Mines Télécom Atlantique\\
  LabSTIC UMR CNRS 6285, UBL\\
  Brest, France \\
  \texttt{ronan.fablet@imt-atlantique.fr} }

\begin{document}

\maketitle

\begin{abstract}
This paper addresses the understanding and characterization of residual networks (ResNet), which are among the state-of-the-art deep learning architectures for a variety of supervised learning problems. We focus on the mapping component of ResNets, which map the embedding space towards a new unknown space where the prediction or classification can be stated according to linear criteria. We show that this mapping component can be regarded as the numerical implementation of continuous flows of diffeomorphisms governed by ordinary differential equations. Especially, ResNets with shared weights are fully characterized as numerical approximation of exponential diffeomorphic operators. We stress both theoretically and numerically the relevance of the enforcement of diffeormorphic properties and the importance of numerical issues to make consistent the continuous formulation and the discretized ResNet implementation. We further discuss the resulting theoretical and  computational insights on ResNet architectures.   
\end{abstract}

\section{Introduction}

Deep learning models are the reference models for a wide range of machine learning problems. Among deep learning (DL) architectures, Residual networks (also called ResNets) have become state-of-the-art ones \cite{He:2016ib,He:2016tq} . Experimental evidences emphasize critical aspects in the specification of these architectures for instance in terms of network depths or combination of elementary layers as well as in their stability and genericity. The understanding and the characterization of ResNet and more widely DL architectures from a theoretical point of view remains a key issue despite recent advances for CNN \cite{Mallat:2016jr}.

Interesting insights on ResNets have recently been presented in
\cite{Ruthotto:2018vb,Haber:2017hk,Weinan:2017kz} from an ordinary/partial differential equation (ODE/PDE) point of view. ResNets are regarded as numerical schemes of differential equations. Especially, in \cite{Ruthotto:2018vb}, this PDE-driven setting stresses the importance of numerical stability issues depending on the selected ResNet configuration. Interestingly, it makes explicit the interpretation of the ResNet architecture as a depth-related evolution of an input space towards a new space where the prediction of the expected output (for instance classes) is solved  according to a linear operator. This interpretation is also pointed out in \cite{Hauser:2017ta} and discussed in terms of Riemannian geometry.   

In this work, we deepen this analogy between ResNets and deformation flows to relate ResNet and registration problems \cite{Sotiras:hk}, especially diffeomorphic registration \cite{Trouve:1998cxa,Beg:2005gr,Ashburner:2007gp,Arsigny:2006uj}. Our contribution is three-fold: (i) we restate ResNet learning as the learning of a continuous and integral diffeomorphic operator and investigate different solutions, especially exponential operator of velocity fields \cite{Arsigny:2006uj}, to enforce diffeomorphic properties; (ii) we make explicit the interpretation of ResNets as numerical approximations of the underlying continuous diffeomorphic setting governed by ordinary differential equations (ODE); (iii) we provide theoretical and computational insights on the specification of ResNets and on their properties.

This paper is organized as follows. Section \ref{s: problem} relates ResNets to diffeomorphic registrations. We introduce in Section \ref{s: math} the proposed diffeomorphism-based learning framework. Section \ref{s: exp} reports experiments. Our key contributions are further discussed in Section \ref{s: related work}.

\section{From ResNets to diffeomorphic registrations}
\label{s: problem}

ResNets \cite{He:2016ib,He:2016tq} have become state-of-the-art deep learning architectures for a variety of issues, including for instance image recognition \cite{He:2016ib} or super-resolution \cite{Kim:2015wv}. This architecture has been proposed in order to explore performances of very deep models, without training degradation accuracy when adding layers. ResNets proved to be easier to optimize and made it possible to learn very deep models (up to hundreds layers).


\begin{figure}
  \centering
  \includegraphics[width=0.4\textwidth]{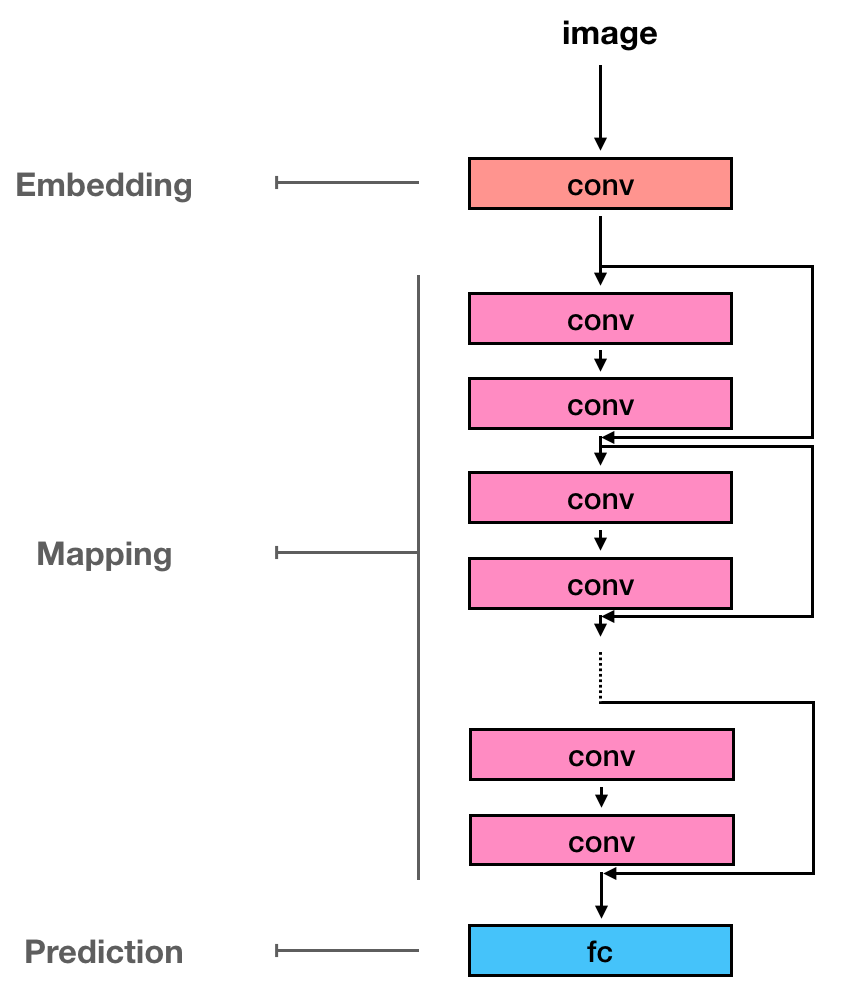}
  \caption{A schematic view of ResNet architecture \cite{He:2016ib}, decomposed into three blocks: embedding, mapping and prediction. 'conv' means convolution operations followed by non linear activations, and 'fc' means fully connected layer.}
  \label{resnet_arch}
\end{figure}

As illustrated in Fig.\ref{resnet_arch}, ResNets can be decomposed into three main building blocks: 
\begin{itemize}
\item the embedding block which aims to extract relevant features from the input variables for the targeted task (such as classification or regression). In \cite{He:2016ib}, the block consists in a set of 64 convolution filters of size $7 \times 7$ followed by non-linear activation function such as ReLU.  
\item the mapping block, which aims to incrementally map the embedding space to a new unknown space, 
in which the data are, for instance, linearly separable in the classification case. In \cite{He:2016ib}, this block consists in a series of residual units. A residual unit is defined as $\mathbf{y} = F(\mathbf{x},\{W_i\}) + \mathbf{x}$ where the function $F$ is the residual mapping to be learned. In \cite{He:2016ib}, $F(\mathbf{x})=W_2 \sigma (W_1 \mathbf{x})$ where $\sigma$ denotes the activation function (bias are omitted for simplifying notations). The operation $F(\mathbf{x}) + \mathbf{x}$ is performed by a shortcut connection and element-wise addition.   
\item the prediction block, which addresses the classification or regression steps from the mapped space to the output space. This prediction block is expected to involve linear models. In \cite{He:2016ib}, this step is performed with a fully connected layer.
\end{itemize}


In this work, we focus on the definition and characterization of the mapping block in ResNets. The central idea of ResNets is to learn the additive residual function $F$ such that the layers in the mapping block are related by the following equation:
\begin{equation}
\mathbf{x}_{l+1} = \mathbf{x}_{l} + F(\mathbf{x}_{l}, W_l) 
\end{equation}
where $\mathbf{x}_{l}$ is the input feature to the $l^{th}$ residual unit. $W_l$ is a set of weights (and biases) associated with the $l^{th}$ residual unit. In \cite{He:2016tq}, it appears that such formulation exhibits interesting backward propagation properties. More specifically, it implies that the gradient of a layer does not vanish even when the weights are arbitrarily small.

Here, we relate the incremental mapping defined by these ResNets to diffeomorphic registration models~\cite{Sotiras:hk}. These registration models, especially Large Deformation Diffeomorphic Metric Mapping (LDDMM) \cite{Trouve:1998cxa,Beg:2005gr}, tackle the registration issue from the composition of a series of incremental diffeomorphic mappings, each individual mapping being close to the identity. Conversely, in ResNet architectures, the $l^{th}$ residual block provides an update of the form $\mathbf{x}_{l} + F(\mathbf{x}_{l}, W_l)$. Under the assumption that $\|F(\mathbf{x}_{l}, W_l)\| \ll \|\mathbf{x}_{l}\|$, the deformation flows generated by ResNet architectures may be expected to implement the composition of a series of incremental diffeomorphic mappings.

In \cite{He:2016ib,He:2016tq}, it is mentioned that the form of the residual function $F$ is flexible. Several residual blocks have been proposed and experimentally evaluated such as bottleneck blocks~\cite{He:2016ib} or various shortcut connections \cite{He:2016tq}. However, by making the connection between ResNet and diffeomorphic mapping, we show here that the function $F$ is a parametrization of an elementary deformation flow, constraining the space of admissible residual unit architectures.

We argue this registration-based interpretation motivates the definition of ResNet architectures as the numerical implementation of continuous flows of diffeomorphisms. Section \ref{s: math} details the proposed diffeomorphism-based learning framework in which diffeomorphic flows are governed by ODEs as in the LDDMM setting. 
Interestingly, ResNets with shared weights relate to a particularly interesting case yielding the definition of exponential diffeomorphism subgroups in the underlying Lie algebra.  
Overall, the proposed framework results in: i) a theoretical characterization of the mapping block as an integral diffeomorphic operator governed by an ODE, ii) in considering deformation flows and Jacobian maps for the analysis of ResNets, iii) the derivation of ResNet architectures with additional diffeomorphic constraints.

\section{Diffeomorphism-based learning}
\label{s: math}

\subsection{Diffeomorphisms and driving velocity vector fields}
Registration issues have been widely stated as the estimation of diffeomorphic transformations between input and output spaces, especially in medical imaging~\cite{Sotiras:hk}. Diffeomorphic properties guarantee the invertibility of the transformations, which includes the conservation of topology features. The parameterization of diffeomorphic transformations according to time-varying velocity vector fields has been shown to be very effective in medical imaging \cite{Klein:2009hx}. Beyond its computational performance, this framework embeds the group structure of diffeomorphisms and results in geodesic flows of diffeomorphisms governed by an Ordinary Differential Equation (ODE):
\begin{equation}
\label{eq: non-stat ode}
\frac{d\phi(t)}{dt}= V_t\left( \phi(t) \right)
\displaystyle
\end{equation}
with $\phi(t)$ the diffeomorphism at time $t$, and $V_t$ the velocity vector field at time $t$. $\phi(0)$ is the identity and $\phi(1)$ the registration transformation between embedding space ${\mathcal{X}}$ and output space ${\mathcal{X}}^*$, such that for any element $X$ in ${\mathcal{X}}$ its mapped version in ${\mathcal{X}}^*$ is $\phi(1)(X)$. Given velocity fields $(V_t)_t$, the computation of $\phi(1)(X)$ comes from the numerical integration of the above ODE.

A specific class of diffeomorphisms refers to stationary velocity fields, that is to say velocity fields which do not depend on time ($V_t=V, \forall t$). As introduced in \cite{Arsigny:2006uj}, in this case, the resulting diffeomorphisms define a subgroup structure in the underlying Lie group and yield the definition of the exponential operators. We here only briefly detail these key properties. We let the reader refer to \cite{Arsigny:2006wi} for a detailed and more formal presentation of their mathematical derivation. For a stationary velocity field, the resulting diffeomorphisms belong to the one-parameter subgroup of diffeomorphisms with infinitesimal generator $V$. In particular, they verify the following property: $\forall s,t,  \phi(t) \cdot \phi(s) = \phi(s+t)$, where $\cdot $ stands for the composition operator in the underlying Lie group. This implies for instance that $\phi(1)$ comes to apply $n$ times $\phi(1/2^n)$ for any integer value $n$. Interestingly, this one-parameter subgroup yields the definition of diffeomorphisms $(\phi(t))_t$ as exponentials of velocity field $V$ denoted by $(\exp (tV))_t$ and governed by the stationary ODE  
\begin{equation}
\label{eq: stat ode}
\frac{d\phi(t)}{dt}= V\left( \phi(t) \right)
\displaystyle
\end{equation}
Conversely, any one-parameter subgroup of diffeomorphisms is governed by an ODE with a stationary velocity  field. It may be noted that the above definition of exponentials of velocity fields generalizes the definition of exponential operators for matrices and finite-dimensional spaces. 

\subsection{Diffeomorphism-based supervised learning}

In this section, we view supervised learning issues as the learning of diffeomorphisms according to some predefined loss function. Let us consider a typical supervised classification issue which the goal is to predict a class $Y$ from an $N$-dimensional real-valued observation $X$. Let ${\mathcal{L}}_\theta$ be a linear classifier model with parameter $\theta$. Within a neural network setting, ${\mathcal{L}}_\theta$ typically refers to a fully-connected layer with softmax activations and parameter vector $\theta$ to the weight and bias parameters of this layer. Let ${\mathcal{D}}$ be the group of diffeomorphisms in ${\mathcal{R}}^N$. We state the supervised learning as the joint estimation of a diffeomorphism $\phi \in {\mathcal{D}}$ and linear classification model ${\mathcal{L}}_\theta$ according to:
\begin{equation}
\widehat{\phi},\widehat{\theta}= \arg \min_{\phi, \theta  }  loss \left ( \{ {\mathcal{L}}_\theta \left ( \phi \left (X_i \right ) \right ),Y_i\}_i \right ) 
\end{equation}
with $\{ X_i ,Y_i\}_i$ the considered training dataset and $loss$ an appropriate loss function, typically a cross entropy criterion. Considering the ODE-based parametrization of diffeomorphisms, the above minimization leads to an equivalent estimation of velocity field sequence $(V_t)$ 
\begin{equation}
\label{eq: diff learning non-stationary}
\widehat{(V_t)},\widehat{\theta}= \arg \min_{(V_t), \theta  }  loss \left ( \{ {\mathcal{L}}_\theta \left ( \phi(1) \left (X_i \right ) \right ),Y_i\}_i \right ) \\
\end{equation}
\begin{equation}
\mbox{subject to} 
\left \{\begin{array}{ccl}
\displaystyle \frac{d\phi(t)}{dt} &=& V_t\left( \phi(t) \right)\\~\\
\phi(0)&=&I\\
\end{array}\right.
\end{equation}
When considering stationary velocity fields~\cite{Arsigny:2006uj,Ashburner:2007gp}, this minimization simplifies as
\begin{equation}
\label{eq: learning non-stationary}
\widehat{V},\widehat{\theta}= \arg \min_{(V_t), \theta  }  loss \left ( \{ {\mathcal{L}}_\theta \left ( \exp (V) \left (X_i \right ) \right ),Y_i\}_i \right ) \\
\end{equation}
We may point out that this formulation differs from the image registration problem in the considered loss function. Whereas image registration usually involves the minimization of the prediction error $Y_i-\phi(1) \left (X_i \right )$ with any pair $X_i,Y_i \in {\mathcal{R}}^N$, we here state the inference of the registration operator $\phi(1)$ according to classification-based loss function. It may also be noted that the extension to other loss functions is straightforward.

 \subsection{Derived NN architecture}
 
To solve for minimization issues (\ref{eq: diff learning non-stationary}) and (\ref{eq: learning non-stationary}), additional priors on the velocity fields can be considered. One may consider the introduction of an additional term in the minimization, which typically involves the integral of the norm of the gradient of the velocity fields and favours small registration displacements between two time steps \cite{Beg:2005gr, Younes:2010eu}. Parametric priors may also be considred. They come to set some parameterization for the velocity fields. In image registration studies, spline-based parameterization has for instance been explored \cite{Ashburner:2007gp}.   


Here, we combine these two types of priors. We exploit a parametric approach and consider neural-network based representations of the driving velocity fields in ODEs (\ref{eq: non-stat ode}) and 
(\ref{eq: stat ode}). More specifically, the discrete parametrization of the velocity field, $V_t(\mathbf{x})$, can be considered as a linear combination of basis functions:
\begin{equation}
\label{eq: param V}
V_t(\mathbf{x}) = \sum_i \nu_{t,i} f_{t,i}(\mathbf{x})
\end{equation}
where $\nu_{t,i}$ are weighting coefficients and $f_{t,i}$ is the $i^{th}$ basis function at time $t$. In this work, $f_{t,i}(\mathbf{x}) = \sigma(W_{l,i} \mathbf{x})$ and corresponds to the $l^{th}$ 2-layer residual unit. Various types of shortcut connections and various usages of activations experimented in \cite{He:2016tq} correspond to various forms of the parametrization of the velocity field. Understanding residual units in a registration-based framework allows to provide a methodological guide to propose new valid residual units. For instance, it has been noticed that adding an activation function such as ReLU after the shortcut connection (\textit{i.e.} after the addition layer) as in \cite{He:2016ib} makes the mapping no more bijective, and thus such architecture may be less efficient, as shown experimentally in \cite{He:2016tq}.




In the registration-based framework considered so far, the transformation $\phi$ is only applied to the observation $X$. This can introduce an undesirable asymmetry in the optimization process and have a significant impact on the registration performance. Inverse consistency, first introduced by Thirion in \cite{Thirion:um}, can be performed by adding a variational penalty term. In order to implement inverse consistent algorithms, it is useful to be able to integrate backwards as well as forwards. In the diffeomorphic framework, the inverse consistency can be written as follows:
\begin{equation}
\phi(1) \circ \phi(-1) = \phi(-1) \circ \phi(1) = \phi(0) 
\end{equation}
This inverse consistency can then be achieved by adding the following term in the overall loss function: 
\begin{equation}
\widehat{\phi},\widehat{\theta}= \arg \min_{\phi, \theta  }  loss \left ( \{ {\mathcal{L}}_\theta \left ( \phi \left (X_i \right ) \right ),Y_i\}_i \right ) + \lambda \sum_i\left(X_i - \phi(-1)(X^*_i) \right)^2
\end{equation}
where $X^*_i = \phi(1)(X_i)$, $X_i \in \mathcal{X}$ and $\lambda$ is a weighting parameter. We may stress that this term does not depend on the targeted task (\textit{i.e.} classification or regression) and only constraint the learning of the mapping block. Thus, this regularization term can be extended to data points that do not belong to the learning set, and more generally to points in a given domain, such that the inverse consistency property does not depend on the sampling of the learning dataset. 

\section{Experiments}
\label{s: exp}

\subsection{Experimental setting}
In this work, following the work on differential geometry analysis of ResNet architectures of Hauser \textit{et al.} in \cite{Hauser:2017ta}, we consider a classification task of 2-dimensional spiral data. The purpose of the mapping block is to warp the input data points $X_i$ into an unknown space $\mathcal{X}^*$ where the transformed data $X^*$ are linearly separable. We have considered the following setting: the loss function is the binary cross-entropy between the output of a sigmoid function applied to the transformed data points $X^*$ and the true labels. Each network is composed of $20$ residual units for which nonlinearities are modeled with $tanh$ activation functions and $10$ basis functions are used for the parametrization of the velocity fields. Weights are initialized with the Glorot uniform initializer (also called Xavier uniform initializer) \cite{Glorot:2010uc}. We use $\ell_2$ weight-decay regularization set to $10^{-4}$ and ADAM optimization method \cite{Kingma:2014us} with a learning rate of $0.001$, $\beta_1=0.9$, $\beta_2=0.999$, minibatch of $300$, $1000$ epochs.

We consider four ResNet architectures: a) a ResNet without shared weights (corresponding to time-varying velocity fields modeling), b) ResNet with shared weights (corresponding to the stationary velocity fields modeling), c) Data-driven Symmetric ResNet with shared weights (considering also the inverse consistency criterion is computed over training data) and d) Domain-driven Symmetric ResNet with shared weights (where the inverse consistency criterion is computed over the entire domain using a random sampling scheme). Although all methods achieved very high classification rates, it can be seen that adding constraints such as the use of stationary velocity fields (\textit{i.e.} share weights) and inverse consistency constraints lead to smoother decision boundaries with no effect on the overall accuracy.
\subsection{Characterization of ResNet properties}

ResNet architectures have been recently studied from the point of view of differential geometry in \cite{Hauser:2017ta}. In this article, Hauser \textit{et al.} have studied the impact of residual-based approaches (compared to non-residual networks) in term of differentiable coordinate transformations. In our work, we propose to go one step further by considering the characterization of the estimated deformation fields leading to an adapted configuration for the considered classification task. More specifically, we consider in this work the maps of Jacobian values.

The Jacobian (\textit{i.e.} the determinant of the Jacobian matrices of the deformations) is defined in a 2-dimensional space as follows:
\begin{equation}
J_{\phi}(\mathbf{x}) = \begin{vmatrix} 
\frac{\partial \phi_1 (\mathbf{x})}{\partial x_1} & \frac{\partial \phi_1 (\mathbf{x})}{\partial x_2} \\ 
\frac{\partial \phi_2 (\mathbf{x})}{\partial x_1} & \frac{\partial \phi_2 (\mathbf{x})}{\partial x_2} \\ 
\end{vmatrix}
\end{equation}

From a physical point of view, the value of the Jacobian represents the local volume variation induced by the transformation. A transformation with a Jacobian value equal to 1 is a transformation that preserves volumes. A Jacobian value greater than 1 corresponds to an expansion and a value less than 1 corresponds to a contraction. The case where the Jacobian is zero means that several points are warped onto a single point: this case corresponds to the limit case from which the bijectivity of the transformation is not verified any more, thus justifying the constraint on the positivity of the Jacobian in several registration methods \cite{Sotiras:hk}.

\subsection{Results}

\begin{figure}[htb]
  \centering
  \includegraphics[width=1.0\textwidth]{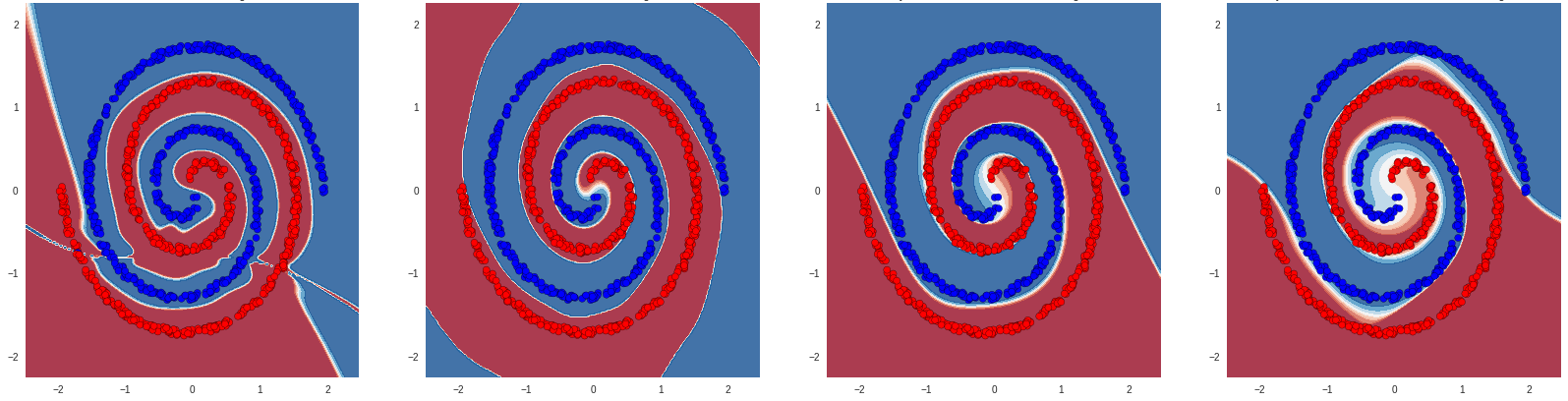}
  \caption{Decision boundaries for the classification task of 2-dimensional spiral data. From left to right: ResNet without shared weights, ResNet with shared weights, Data-driven Symmetric ResNet with shared weights, Domain-driven Symmetric ResNet with shared weights. We refer the reader to the main text for the correspondence between ResNet architectures and diffeomorphic flows.}
  \label{fig:boundary}
\end{figure}

Classification algorithms are usually only evaluated using classification accuracy (as the number of correct predictions from all predictions made). However, classification rate is not enough to characterize performances of specific algorithm. 
In all the experiments shown in this work, the classification rate is greater than $99\%$. Visualization of the decision boundary is an alternative way to provide complementary insights on the regularity of the solution in the embedding space. Fig.~\ref{fig:boundary} shows the decision boundary for the four considered ResNets. Although all methods achieved very high classification rates, it can be seen that adding constraints such as the use of stationary velocity fields (\textit{i.e.} shared weights) and inverse consistency constraints lead to smoother decision boundaries with no effect on the overall accuracy. This is regarded as critical for generalizability and adversarial robustness \cite{Szegedy:2013vw}. 

\begin{figure}[htb]
  \centering
  \includegraphics[width=1.0\textwidth]{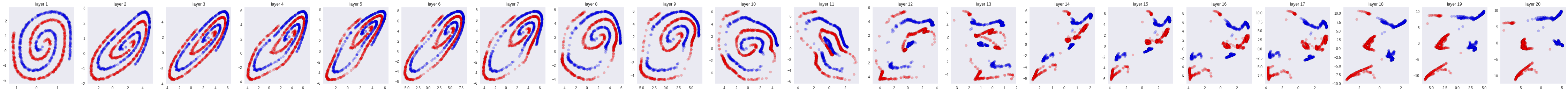}
  \includegraphics[width=1.0\textwidth]{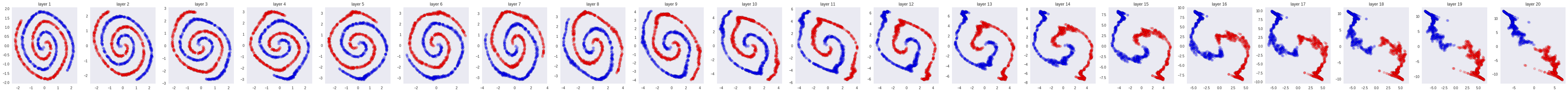}
  \includegraphics[width=1.0\textwidth]{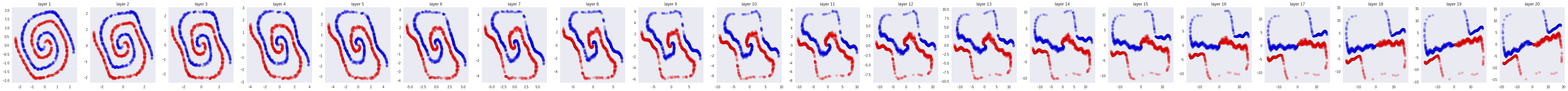}
  \includegraphics[width=1.0\textwidth]{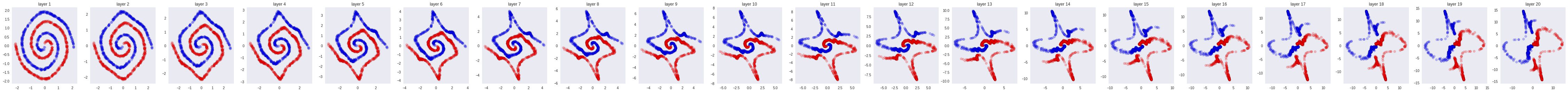}
  \caption{Evolution of the spatial configuration of data points through the 20 residual units. From top to bottom: ResNet without shared weights, ResNet with shared weights, Data-driven Symmetric ResNet with shared weights, Domain-driven Symmetric ResNet with shared weights.}
  \label{fig:motion}
\end{figure}

Decision boundaries correspond to the projection of the estimated linear decision boundary in the space $\mathcal{X}^*$ into the embedding space $\mathcal{X}$. The visualization of decision boundaries does not however provide information regarding the topology of the manifold in the output space $\mathcal{X}^*$. We also study the deformation flow trough the spatial configuration of data points through the network layers as in \cite{Hauser:2017ta}. Figure\ref{fig:motion} shows how each network untangles the spiral data. Networks with shared weights exhibit smoother layer-wise transformations. More specifically, this visualization provides insights on the geometrical properties (such as topology preservation / connectedness) of the transformed set of input data points. 

\begin{figure}[htb]
  \centering
  \includegraphics[width=1.0\textwidth]{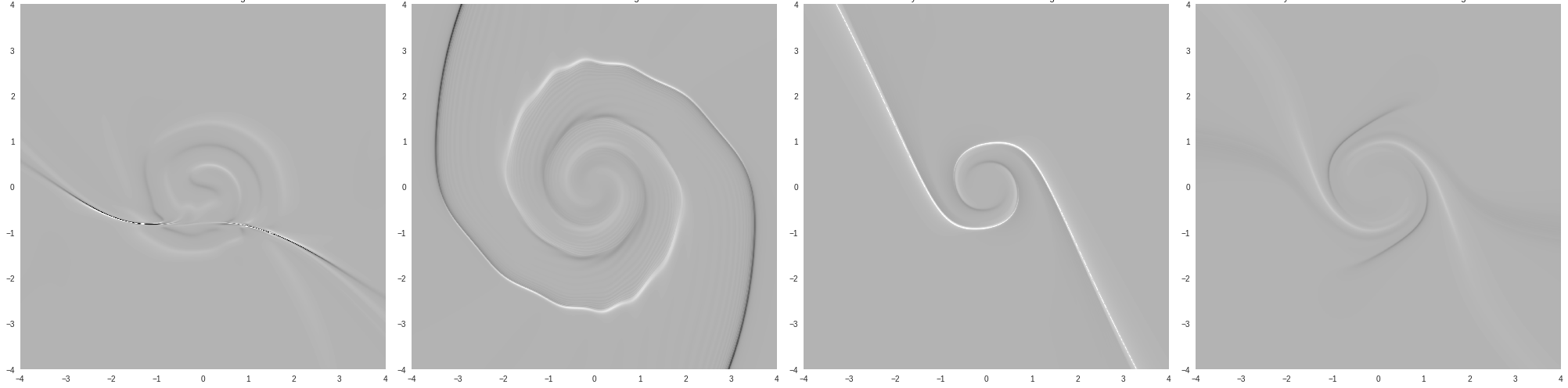}
  \caption{Jacobian maps for the four ResNet architectures. From left to right: ResNet without shared weights ($J_{min}=-5.59$, $J_{max}=6.34$), ResNet with shared weights ($J_{min}=-1.41$, $J_{max}=2.27$), Data-driven Symmetric ResNet with shared weights ($J_{min}=0.55$, $J_{max}=5.92$), Domain-driven Symmetric ResNet with shared weights ($J_{min}=0.30$, $J_{max}=1.44$). (colormap : $J_{min}=-2.5$, $J_{max}=2.5$, so dark pixels correspond to negative jacobian values).}
  \label{fig:jacobian}
\end{figure}

To evaluate the quality of the estimation warping transformation, Fig.\ref{fig:jacobian} shows the Jacobian maps for each considered network. Negative jacobian values correspond to locations where bijectivity is not satisfied. It can be seen that adding constraints such as stationary velocity fields and inverse consistency leads to more regular geometrical shapes of the deformed manifold. The domain-driven regularization applied to a ResNet with shared weights leads to the most regular geometrical pattern. 

\section{Discussion: Insights on ResNet architectures from a diffeomorphic viewpoint}
\label{s: related work}

As illustrated in the previous section, the proposed diffeomorphic formulation of ResNets provide new theoretical and computational insights for their interpretation and characterization as discussed below. 

\subsection{Theoretical characterization of ResNet architectures}
In this work, we make explicit the interpretation of the mapping block of ResNet architectures as a discretized numerical implementation of a continuous diffeomorphic registration operator. This operator is stated as an integral operator associated with an ODE governed by velocity fields. Importantly, ResNet architectures with shared weights are viewed as the numerical implementation of exponential of velocity fields, equivalently defined as diffeomorphic operators governed by stationary velocity fields. Exponentials of velocity fields are by construction diffeomorphic under smoothness constraints on the generating velocity fields. Up to the choice of the ODE solver implemented by ResNet architecture (in our case an Euler scheme), ResNet architectures with shared weights are then fully characterized from a mathematical point of view.

The diffeomorphic property naturally arises as a critical property in registration problems, as it relates to invertibility properties. Such invertibility properties are also at the core of the definition of kernel approaches, which implicitly defines mapping operators \cite{Scholkopf:2002ta}. As illustrated for the reported classification experiments, the diffeomorphic property prevents the mapping operator from modifying the topology of the manifold structure of the input data. When not imposing such properties, for instance in unconstrained ResNet architectures as well as, the learned deformation flows may present unexpected topology changes.  

The diffeomorphic property may be regarded as a regularization criterion on the mapping operator, so that the learned mapping enables a linear separation of the classes while guaranteeing the smoothness of the classification boundary and of the underlying deformation flow. It is obvious that a ResNet architecture with shared weights is a special case of an unconstrained ResNet. Therefore, the training of a ResNet architecture with shared weights may be viewed as the training of an unconstrained ResNet within a reduced search space. The same holds for the symmetry property which further constrains the search space during training. The later constraint is shown to be numerically important so that the discretized scheme complies with the theoretical diffeomorphic properties of exponentials of velocity fields.  

Overall, this analysis stresses that over an infinity of mapping operators reaching optimal training performance one may favor those depicting diffeomorphic properties so that key properties such as generalization performance, prediction stability and robustness to adversarial examples are greatly improved. Numerical schemes which fulfill such diffeomorphic properties during the training process could be further investigated and could benefit from the registration literature, including for diffeomorphics flows governed by non-stationary velocity fields \cite{Trouve:1998cxa,Beg:2005gr,Avants:2004io}.

\subsection{Computational issues}

Besides theoretical aspects, computational properties also derive from the proposed diffeomorphism-based formulation. Within this continuous setting, the depth of the network relates to the integration time step and the precision of the integration scheme. The deeper the network, the smaller the integration step. Especially, a large integration time step, {\em i.e.} a shallower ResNet architecture, may result in numerical integration instabilities and hence in non-diffeomorphic transformations 
Therefore, deep enough architectures should be considered to guarantee numerical stability and diffeomorphic properties. The maximal integration step relates to the regularity of the velocity fields governing the ODEs. In our experiments, we only consider an explicit first-order Euler scheme. Higher-order explicit schemes, for instance the classic fourth-order Runge-Kutta scheme, seem of great interest as well as implicit integration schemes \cite{Davis:1984vu}. Given the spatial variabilities of the governing velocity fields, adaptive integration schemes also appear as particularly relevant. 

Diffeomorphic mapping defined as exponential of velocity fields were shown to be computationally more stable with smoother integral mappings. They lead to ResNet architectures with shared weights, which greatly lower the computational complexity and memory requirements compared with the classic ResNet architectures. They can be implemented as Recurrent Neural Networks \cite{Kim:2015wa,Liao:2016tj}. Importantly, the NN-based specification of the elementary of velocity field $V$ (\ref{eq: param V}) becomes the bottleneck in terms of modeling complexity. 
The parametrization (Equation~\ref{eq: param V}) may be critical to reach good prediction performance.
Here, we considered a two-layer architecture regarded as a projection of $V$ onto basis function. Higher-complexity architecture, for instance with larger convolution supports, more filters or layers, might be considered while keeping the numerical stability of the overall ResNet architectures. By contrast, considering higher-complexity elementary blocks in a ResNet architectures without shared weights would increase numerical instabilities and may required complementary regularization constraints across network depth \cite{He:2016ib,Ruthotto:2018vb}.  

Regarding training issues, our experiments exploited a classic backpropagation implementation with a random initialization. From the considered continuous log-Euclidean prospective, the training may be regarded as the projection of the random initialization onto the manifold of acceptable solutions, {\em i.e.} solutions satisfying both the minimization of the training loss and diffeomorphic constraints. In the registration literature \cite{Sotiras:hk}, the numerical schemes considered for the inference of the mapping usually combine a parametric representation of the velocity fields and a multiscale optimization strategy in space and time. The combination of such multiscale optimization strategy to backpropagation schemes appears as a promising path to improve convergence properties, especially the robustness to the initialization. The different solutions proposed to enforce diffeomorphic properties are also of interest. Here, we focused on the invertibility constraints, which result in additional terms to be minimized in the training loss. 





\section{Conclusion}

This paper introduces a novel registration-based formulation of ResNets. We provide a theoretical interpretation of ResNets as numerical implementations of continuous flows of diffeomorphisms. Numerical experiments support the relevance of this interpretation, especially the importance of the enforcement of diffeomorphic properties, which ensure the stability and generalization properties of a trained ResNet. This work opens new research avenues to explore further diffeomorphism-based  formulations and associated numerical tools for ResNet-based learning, especially regarding numerical issues.

\subsubsection*{Acknowledgments}

We thank B. Chapron, L. Drumetz, and C. Herzet for their comments and suggestions. The research leading to these results has been supported by the ANR MAIA project, grant ANR-15-CE23-0009 of the French National Research Agency, INSERM and Institut Mines Télécom Atlantique (Chaire ”Imagerie médicale en thérapie interventionnelle”) and Fondation pour la Recherche Médicale (FRM grant DIC20161236453), Labex Cominlabs (grant SEACS), CNES (grant OSTST-MANATEE) and Microsoft trough AI-for-Earth EU Oceans Grant (AI4Ocean). We also gratefully acknowledge the support of NVIDIA Corporation with the donation of the Titan Xp GPU used for this research.

\bibliographystyle{abbrv}
\bibliography{biblio}

\end{document}